\documentclass[conference]{IEEEtran}
\IEEEoverridecommandlockouts
\usepackage{cite}
\usepackage{amsmath,amssymb,amsfonts}
\usepackage{algorithmic}
\usepackage{graphicx}
\usepackage{textcomp}
\usepackage{xcolor}
\usepackage{array}
\usepackage{url}
\usepackage{hyperref}
\usepackage{cleveref}
\usepackage{adjustbox}
\usepackage{float}
\usepackage{comment}	
\usepackage{siunitx}    

\usepackage{tikz}
\usepackage{textcomp}
\usepackage{lipsum}

\newcommand\copyrighttext{%
	\footnotesize \textcopyright ©2020 IEEE. Personal use of this material is permitted. Permission from IEEE must be obtained for all other uses, in any current or future media, including reprinting/republishing this material for advertising or promotional purposes, creating new collective works, for resale or redistribution to servers or lists, or reuse of any copyrighted component of this work in other works.
}
\newcommand\copyrightnotice{%
	\begin{tikzpicture}[remember picture,overlay]
	\node[anchor=south,yshift=10pt] at (current page.south) {\fbox{\parbox{\dimexpr\textwidth-\fboxsep-\fboxrule\relax}{\copyrighttext}}};
	\end{tikzpicture}%
}

\def\BibTeX{{\rm B\kern-.05em{\sc i\kern-.025em b}\kern-.08em
    T\kern-.1667em\lower.7ex\hbox{E}\kern-.125emX}}
\begin{document}

\title{Lane-Change Initiation and Planning Approach for Highly Automated Driving on Freeways\\
\thanks{This work was supported by Jaguar Land Rover and the UK-EPSRC grant EP/N01300X/1 as part of the jointly funded Towards Autonomy: Smart and Connected Control (TASCC) Programme.}
\thanks{\IEEEauthorrefmark{2}All work was done while the author was an internship student at the University of Surrey.}
}

\author{
\IEEEauthorblockN{Salar Arbabi\IEEEauthorrefmark{1},
Shilp Dixit\IEEEauthorrefmark{1},
Ziyao Zheng\IEEEauthorrefmark{2},
David Oxtoby\IEEEauthorrefmark{3},
Alexandros Mouzakitis\IEEEauthorrefmark{3} and
Saber Fallah\IEEEauthorrefmark{1}}
\IEEEauthorblockA{\IEEEauthorrefmark{1}Department of Mechanical Engineering Sciences, University of Surrey, Guildford, GU2 7XH, UK. 
\IEEEauthorblockA{\IEEEauthorrefmark{2}Électronique et Technologies Numériques(ETN), École Polytechnique de l'Université de Nantes, 44306 Nantes, France.}
\IEEEauthorblockA{\IEEEauthorrefmark{3}Jaguar Land Rover Limited, Coventry, CV3 4LF, UK.}
}}
\maketitle
\copyrightnotice
\begin{abstract}
Quantifying and encoding occupants' preferences as an objective function for the tactical decision making of autonomous vehicles is a challenging task.
This paper presents a low-complexity approach for lane-change initiation and planning to facilitate highly automated driving on freeways. Conditions under which human drivers find different manoeuvres desirable are learned from naturalistic driving data, eliminating the need for an engineered objective function and incorporation of expert knowledge in form of rules.
Motion planning is formulated as a finite-horizon optimisation problem with safety constraints. 
It is shown that the decision model can replicate human drivers' discretionary lane-change decisions with up to 92\% accuracy. Further proof of concept simulation of an overtaking manoeuvre is shown, whereby the actions of the simulated vehicle are logged while the dynamic environment evolves as per ground truth data recordings.

\end{abstract}

\begin{IEEEkeywords}
Autonomous Vehicles, Decision Making, Motion Planning
\end{IEEEkeywords}

\section{Introduction}
Advanced Driver Assistance Systems (ADAS) and automated highway driving are expected to improve passenger's safety and comfort \cite{Bengler2014}. Every year, new systems are offering a higher degree of autonomous functionality. For example, Audi introduced the A8 model which is equipped with the traffic jam pilot, capable of performing simple manoeuvres such as lane and distance keeping. Tesla's Autopilot system is now equipped with a lane-changing feature that can initiate and perform lane changes autonomously. 

\begin{figure}[t!]
	\centering
	\includegraphics[width=0.95\linewidth]{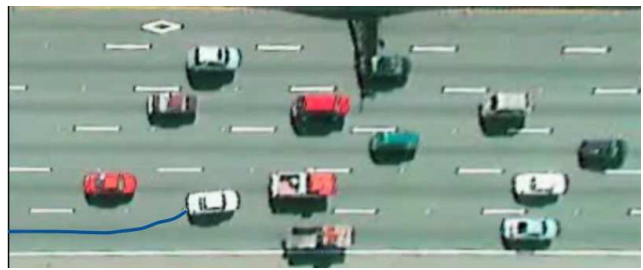}
	\caption{An example scenario where a driver has decided to overtake the slow-moving truck on the innermost lane. The image has been taken from the NGSim dataset that is used for training the decision model.}
	\label{fig:ngsim}
\end{figure}

In \autoref{fig:ngsim}, a typical scenario in which a driver intents to circumvent a slow-moving vehicle is shown. 
Several factors, such as the length of time the situation has persisted and the availability of gaps in the adjacent lane govern the tactical decision-making of the driver. Similar contextual information can be leveraged by autonomous cars to handle various traffic scenarios.  
Often, the decision systems rely on functions that are hand-crafted and can be tedious and time consuming to tune \cite{Paden2016}. In our previous work \cite{dixit2018trajectory}, preferences and traffic rules were embedded in several potential functions, superposition of which resulted in a potential field. The potential field was combined with reachable sets to guide the vehicle towards safe zones on the road while avoiding collision with other vehicles. Hence, the emerging behaviour of the vehicle was fundamentally rule-based. Approaches based on machine learning, namely Reinforcement Learning (RL) \cite{hoel2019combining, kuutti2019end} and Supervised Learning (SL) \cite{Vallon2017, eraqi2017end} alleviate some of these limitations and offer greater ability to generalize to various situations a vehicle may encounter. However, these learning approaches suffer from two key drawbacks of their own: the need for high fidelity simulators for RL, and the need for large scale collection and annotation of data for SL, which is often costly. 

In this paper,
\begin{itemize}
	\item We utilise naturalistic vehicle trajectories captured from CCTV camera recordings as training data for learning a decision model (\Cref{sec:data_prep}). The decision model maps the driving context at each time instance to two manoeuvre classes: discretionary lane change and lane-keeping.
	
	\item We propose a set of features and techniques to improve the accuracy of the decision model (\Cref{sec:features} and \Cref{sec:temporal}).
	
	\item We build on the model predictive control (MPC) framework presented in previous work\cite{dixit2018trajectory} to turn the selected manoeuvre by the decision model into continuous trajectories while satisfying safety and comfort constraints (\Cref{sec:trajectoryplanning}).
\end{itemize}

\section{Related Work} \label{sec:relatedwork}
\label{sec:}
There has been extensive work on motion planning for self-driving cars, with the objective of generating collision-free trajectories while minimizing jerky motion \cite{Paden2016}. For tactical decision making, early-stage autonomous vehicles were programmed to react to their immediate situation based on context parameters such as time to collision \cite{Urmson2009,Montemerlo2009}. 
Ulbrich and Maurer \cite{Ulbrich2015b} proposed a decision-making framework for executing lane changes, while also addressing uncertainties that result from occlusions. For behaviour planning, authors employ Partially Observable Markov Decision Processes (POMDP) with tree-based policy evaluation for selecting a sequence of actions that result in the highest utility. Ardelt et al. \cite{Ardelt2012} presented BMW's hierarchical decision-making framework. Highway driving was divided into several discrete behaviours, which are chosen at each time instance using criteria such as the preferred driving speed and the gap size available in the adjacent lane. 
All these approaches incorporate domain knowledge in the action selection process, which can be hard to encode. Furthermore, the resulting behaviour of the autonomous car can become dependent on subjective design choices and parameter tunings.

Recently, deep reinforcement learning has been proposed for tactical decision-making \cite{hoel2019combining}. Although the approach has shown promising results, for implementation on real systems, high fidelity simulators have to be developed for training the learning agent, which poses a challenge of its own. 
Vallon et al.\cite{Vallon2017} proposed a decision model based on the support vector machine (SVM) classifier to determine when a lane change should be initiated. The classifier was trained on collected data from both defensive and aggressive driving styles to create a more personalised lane change experience. It appears that our work and \cite{Vallon2017} share the same goal of exploiting learned decision models for autonomous driving on freeways. However, in \cite{Vallon2017}, authors collected their driving data using a test vehicle, with a total of 25 lane change manoeuvres recorded. In our work, by utilising vehicle trajectories captured from CCTV cameras, 600 lane change manoeuvres were automatically extracted based on the vehicles' motion and position on the road. A larger dataset improves the generalization capability of the decision model. Furthermore, in our work, unlike \cite{Vallon2017}, we make use of a set of engineered features that improve the decision model's accuracy.  

\section{SYSTEM OVERVIEW} \label{sec:systemarchitecture}
\begin{figure}[t!]
	\centering
	\includegraphics[width=0.9\linewidth]{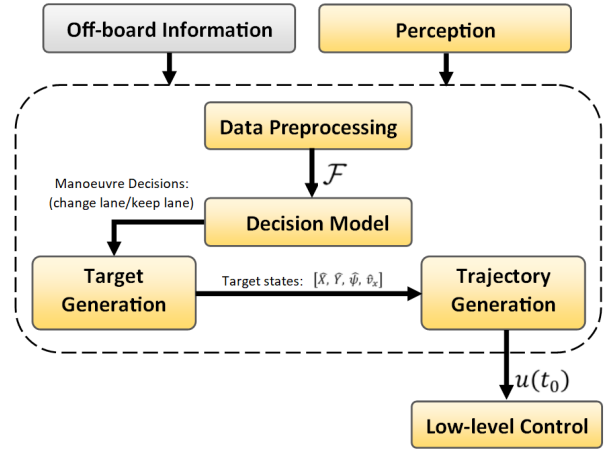}
	\caption{System overview for the proposed approach.}
	\label{fig:architecture}
\end{figure}

The system overview for the proposed approach is shown in \autoref{fig:architecture}. While the framework consists of multiple modules, our contribution focuses on the
component that is tasked with generating high-level manoeuvres, and motion planning for executing these manoeuvres.  Given the feature vector $\mathcal{F}$, which compactly represents the current driving situation, appropriate manoeuvres are determined by the decision model. 
Depending on the chosen manoeuvre, Target Generation module identifies the Lead-Vehicle (LV) when one exists. The subject vehicle (SV), i.e., the autonomous vehicle, has to be cognizant of LV and keep a safe distance from it. In the current implementation, LV is always set to be the lead vehicle in SV's desired lane. Nevertheless, the approach could be extended with a motion prediction module such as the one proposed in \cite{bahram2016combined} for assigning LV to vehicles that intent to merge in front of the SV. Once the LV is identified, a reference state set point (i.e., velocity, lateral position, and heading angle) is computed to be tracked by the MPC-based Trajectory Generation module.
If there is no preceding vehicle, a target position is set within the safe zones of a predefined potential field.  
While the framework's modular architecture allows it to be integrated into existing ADAS, it can eventually be extended for fully autonomous driving on freeways.

\section{DECISION MODEL} \label{sec:decisionmodel}
The decision model provides the mapping between the current scene context and two manoeuvre classes: \textit{lane keeping} and \textit{discretionary lane Change} (See \autoref{fig:ngsim} for a motivating example scenario). In this work, the mapping is obtained using Random Forests (RF), a popular machine learning algorithm for classification. 

\subsection{Data Preprocessing and Extraction} \label{sec:data_prep}
For training the decision model, we use trajectories captured from CCTV cameras mounted at two locations: US 101 in Los Angeles, California, and I-80 interstate in the San Francisco bay area, California \cite{NGSIM}. Since the raw data is noisy, the extended Kalman filter with a bicycle model and the exponential smoothing algorithm in\cite{NGSIM.jl} were used for obtaining a more accurate estimate of vehicles' positions and velocities.

\begin{figure} 
	\centering
	\includegraphics[width=0.98\linewidth]{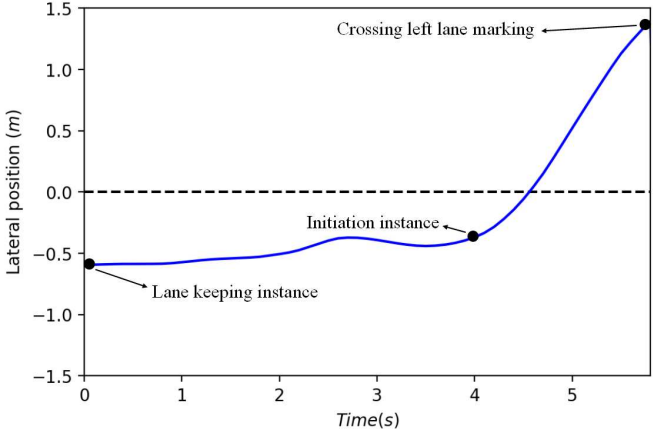}
	\caption{An example lateral motion profile of a vehicle from the dataset. Lane change initiation is defined as the instance at which the lateral speed exceeds \SI{0.1}{\meter\per\second}.}
	\label{fig:lc_detect}
\end{figure}

Trajectories were partitioned into two sets, one for lane-keeping and one for the lane change manoeuvre. The way the sets were labelled is as follows. All the time instances at which a driver changes lane were gathered first. These were indicated by the time instance at which the front of a vehicle crosses the lane marking (See \autoref{fig:lc_detect}). For each lane-change manoeuvre, the initiation point is labelled as the time when the lateral speed of a vehicle exceeds \SI{0.1}{\meter\per\second}. An equal number of lane-keeping instances were sampled from the period before the initiation point, resulting in a balanced training set. However, this approach comes at the cost of limiting the model's exposure to many instances in which a lane-change is not desirable or acceptable. This is an inherent problem since most of highway driving does not involve performing lane changes. Increasing the size of the dataset is one solution for overcoming this problem.

\begin{table}
	\caption{Extracted features}
	\begin{center}
		\begin{adjustbox}{max width=0.98\linewidth}
			\begin{tabular}{>{\raggedright}p{1cm}>{\raggedright\arraybackslash}p{7.5cm}}
				
				\textbf{Features} & \textbf{Description}\\
				\hline
				
				$x_{_{RL}}$ & Longitudinal distance between the rear left vehicle and the SV \\ 
				$TTC_{_{RL}}$ & Time to collision with rear left vehicle \\ 
				$\Delta$ & Utility feature - a model of drivers' contentment\\
				$TTC_{_{FL}}$ & Time to collision with front left vehicle \\ 
				$v_{_{FL}}$ & Relative speed between the SV and front left vehicle   \\ 
				$x_{_{FL}}$ & Longitudinal distance between the front left vehicle and the SV  \\
				$v_{_{RL}}$ & Relative speed between the rear left vehicle and the SV  \\
				$v_{_{LV}}$ & Relative speed between SV and LV  \\
				$x_{_{LV}}$ & Longitudinal distance between the LV and the SV   \\
				$vel{_{SV}}$ & SV's speed  \\
				
				\label{table:features}
			\end{tabular}
		\end{adjustbox}
	\end{center}
\end{table}
\subsection{Training Features} \label{sec:features}
A set of features which encode semantically meaningful information for representing a driver's local context were extracted from the dataset. 
The feature vector contains, for example, the relative positions and velocities between an ego car and its surrounding vehicles, which are generally available to an autonomous driving system. List of all the used features and their descriptions are provided in \autoref{table:features}.
Additionally, we make use of a utility feature  $\Delta$, expressed as,
\begin{equation} \label{equ:utility}
\begin{aligned} 
\text{$\Delta =$}
\sum_{t=0}^{T}\;\frac{{v_{LV}}^{(t)}}{{x_{LV}}^{(t)}}
\end{aligned}
\end{equation}
where ${v_{LV}}^{(t)}$ and ${x_{LV}}^{(t)}$ are the relative speed and distance between the ego driver and the LV at time $t$, respectively. The purpose of the utility feature is to model the drivers' level of contentment with a given driving situation. The summation expresses the intuition that dissatisfaction associated with driving behind a slower car increases while the situation persists.
Time $t$ is reset every time the LV changes. The relative speed has a negative magnitude when the distance between the two vehicles shrinks. The utility at each instance is inversely proportional to ${x_{LV}}^{(t)}$. This inverse relationship captures the increase in LV's influence on the ego's utility as the distance between the two falls.
If at any time the magnitude of $\Delta$ is positive (i.e implying that the driver is content) while its gradient is negative (i.e. his or her level of contentment is diminishing), value of $\Delta$ is reset to zero,
\begin{equation} \label{equ:loss}
\begin{aligned} 
\text{$\Delta =$}
\begin{cases} 
\Delta, & if \; \Delta < 0   \\
\Delta, & if \; \Delta'' > 0 \;\;\&\;\; \Delta > 0  \\
0, &  if \; \Delta'' < 0 \;\;\&\;\; \Delta > 0 

\end{cases}
\end{aligned}
\end{equation}
In other words, a negative utility is retained overtime, while a positive utility is set to zero with any negative experience. The intuition follows psychological studies, which hypothesize that humans give greater weight to negative experiences than positive ones \cite{rozin2001negativity}. Our use of the utility feature is novel, and we found that it boosts the performance of the decision model. 

Since we are only interested in discretionary lane changes and the true intention of drivers is hidden, we excluded those lane changes from the dataset set that met the following criteria:

\begin{enumerate}
	\item Lane change was to the right lane, on the basis that these were mandatory lane changes.	
	\item Lane change was performed near the highway entrance and exit ramp.
\end{enumerate}
Additionally, we only considered lane changes during which the time headway between the ego, the rear left and the front left vehicles was greater than $\SI{2}{\second}$ at the initiation instance. This was done to stop the model from learning aggressive behaviours. Overall, a total of 600 lane change manoeuvres were extracted.

%

\subsection{Random Forests}
Random Forests  (RF) \cite{breiman2001random} is an ensemble of $\mathcal{T}$ decision trees each assigning a probability to a given class $c$ : $p_{n}(c|\mathcal{F})$. The tree predictors are trained on a set of randomly sampled feature vectors to minimize the residual error in the predictions.
Each tree grows until there is only one class left in each of the leaf nodes. Once trained, at each time instance the RF-based decision model assigns a probability to each manoeuvre class:

\begin{equation} \label{equ:RF}
\begin{aligned} 
p(c|\mathcal{F}) = 
\frac{1}{\mathcal{T}} \sum_{n=1}^{\mathcal{T}}\; p_{n}(c|\mathcal{F})
\end{aligned}
\end{equation}
where $\mathcal{T} = 100$ in this implementation and the feature vector $\mathcal{F}$ consists of the context features described in \Cref{sec:features}.

We chose RF since it offers several benefits. Firstly, as the construction of the trees is non-parametric, model complexity is data-driven. This makes the RF model relatively robust to over-fitting. Additionally, decision trees can be constructed sequentially \cite{saffari2009line}, allowing for further personalization of the decision model to particular drivers based on their disengagements. We also explored logistic regression and SVM but found RF to show superior classification accuracy.

\section{TRAJECTORY GENERATION} \label{sec:trajectoryplanning}
For lane keeping, the SV maintains a time headway of $\SI{2}{\second}$ with the LV. Once the desire to change lane is inferred by the decision model, the target pose for the vehicle is updated as follows: (\textit{i}) the lateral position target ($\hat{y}$) is updated to the centre of the desired lane, (\textit{ii}) the heading-angle target ($\hat{\psi}$) is updated to keep the vehicle aligned with the road's curvature, and (\textit{iii}) the longitudinal velocity target ($\hat{v}$) is modified for maintaining a time headway of $\SI{2}{\second}$ with the LV in SV's desired lane. If no LV exists, $\hat{v}$ is set according to preferences and traffic rules encoded as a predefined potential field. 

For the application of the MPC based trajectory planning algorithm, the plant (SV) dynamics are expressed as a linear time-invariant system and discretised using the forward Euler method with sampling time of $T_{\textrm{s}} = \SI{0.1}{\second}$. The discrete-time system can be expressed in the state-space form:
\begin{equation}
\xi \left( k + 1 \right) = A \xi \left( k \right) + B u \left( k \right)	\label{eq:genStateSpace}
\end{equation}
where $\xi = [y,\psi,v]^{\textrm{T}}$ is the state-vector of the system, $u = [\delta_{\textrm{f}},a_{x}]^{\textrm{T}}$ consisting of the front wheel steer angle $\delta_{\textrm{f}}$ and longitudinal acceleration $a_{x}$ is the input vector of the system, and the matrices $A$ and $B$ are constant. It is assumed that the pair $(A,B)$ is stabilisable and that the states and inputs can be represented using a set notation given by $\xi \in \mathcal{X} \subseteq \mathbb{R}^{3}$, $u \in \mathcal{U} \subseteq \mathbb{R}^{2}$.

As illustrated in \cite{dixit2018trajectory}, given a target steady state $\hat{x}$, the control objective is to find a control action of the form $u(k) = F_{H}( \xi(k),\hat{\xi} )$ such that the states of system \eqref{eq:genStateSpace} is steered as close as possible to the target state $\hat{\xi}$ while fulfilling the state and input constraints. The MPC framework described in \cite{limon2008mpc} can be applied to control the system represented in \eqref{eq:genStateSpace}. The resultant constrained optimisation problem is given as
\begin{equation}
\begin{array}{l}
\mathop {\min }\limits_{U_{k},\theta } V_{H}({U_{k}},\theta ;\xi,\hat{\xi})\\
\quad \quad \quad {\rm{subject}}\;{\rm{to}}\\
\quad \quad \quad \xi(0) = \xi\\
\quad \quad \quad \xi(k) \in \mathcal{X} \\
\quad \quad \quad \xi(k) \in \mathcal{X}_{\textrm{ca}} \\
\quad \quad \quad u(k) \in \mathcal{U} \\
\quad \quad \quad \xi(k + 1) = A \xi \left( k \right) + B u \left( k \right),\quad k = 0,1, \ldots ,H\\
\quad \quad \quad (\xi_{\textrm{ss}},\,u_{\textrm{ss}}) = M_{\rho} \rho \\
\quad \quad \quad (\xi(H),\,\theta ) \in \mathcal{X}_{t}
\end{array}	\label{eq:genOptProblem}
\end{equation}
where $H$ is the prediction horizon of the MPC, $\rho$ is a parameter vector that characterises the subspace of steady-states and inputs, $\mathcal{X}_{\textrm{ca}}$ defines the convex set representing the collision-free zone on the road, and the terminal set $\mathcal{X}_{t}$ is designed as in \cite{limon2008mpc}. The performance index $V_{H}({U_{k}},\theta ;\xi,\hat{\xi})$ is defined as
\begin{align} \label{eq:genCostFunc}
\begin{split}
V_{H}({U_{k}},\theta ;\xi,\hat{\xi}) = &\sum\limits_{k = 0}^N \left[ ||\xi(k) - \xi_{\textrm{ss}}||_{Q}^{2} + ||u(k) - u_{\textrm{ss}}||_{R}^{2} \right] + \\
&||\xi(H) - \xi_{\textrm{ss}}||_{P}^{2} + ||\xi_{\textrm{ss}} - \hat{\xi}||_{T}^{2}
\end{split}
\end{align}
The solution to \eqref{eq:genOptProblem} results in an optimal input sequence $U^{*}(k) = (u^{*}(0; \xi,\hat{\xi}),u^{*}(1; \xi,\hat{\xi}), \ldots ,u^{*}(H-1; \xi,\hat{\xi}) )$, and a parametrised steady-state $\rho^{*} ( \xi, \hat{\xi} )$. The net control action applied on the plant is given as:
\begin{equation}
u_{\textrm{mpc}} = u^{*} (0; \xi,\hat{\xi})	\label{eq:netCtrlAction}
\end{equation}
The prediction horizon has been chosen as $H = 30$, weighting matrices $Q = \text{diag}\left( 10^3,~10^{-2},~10^{-2}\right), ~ R = \text{diag}\left( 10^1,~10^{-2}\right)$, $P$ is the solution to discrete-time algebraic Riccati equation, $T = 10^3 \cdot P$, and the nominal control action $K = \left[660.03,~ 262.3416,~ 45 \right]$, which is computed by solving the LQR problem.

\subsection{Collision Avoidance Constraints}	\label{subsec:collAvoidance}
To ensure that the trajectories generated by the MPC controller are safe, additional collision avoidance constraints represented by a convex polyhedron $\mathcal{X}_{\textrm{ca}}$ are added to the optimisation problem in \eqref{eq:genOptProblem}. The hyperplanes describing the polyhedron $\mathcal{X}_{\textrm{ca}}$ are constructed using inequalities of the form
\begin{equation}
a_{\textrm{ca},i} x + b_{\textrm{ca},i} y + c_{\textrm{ca},i} < 0	\label{eq:collisionAvoidanceConst}
\end{equation}
where $i$ is the $i^{\textrm{th}}$ hyperplane representing an edge of the safe collision-free part of the road, $(a_{\textrm{ca}},b_{\textrm{ca}},c_{\textrm{ca}})$ are the parameters depicting the equation of line in 2D space, and $(x,y)$ are the global coordinates. To identify the collision-free zones on the road and generate the collision avoidance set $\mathcal{X}_{\textrm{ca}}$, the technique proposed in \cite{dixit2019trajectory} is utilised where artificial potential fields are used to obtain a local risk-map for the SV. The addition of these collision-avoidance constraints ensure that the trajectories generated by solving the constrained optimisation problem in \eqref{eq:genOptProblem} are restricted to the safe regions of the road for the combination of any admissible $\xi(0)$ and $\hat{\xi}$. An interested reader is directed to \cite{dixit2019trajectory} for further design details of these constraints.
At each discrete time instant $k$, the problem in \eqref{eq:genOptProblem} with additional constraints \eqref{eq:collisionAvoidanceConst} is solved by setting the target state and the initial state to $\hat{\xi} = [\hat{y},\hat{\psi},\hat{v}]^{\textrm{T}}$ and $\xi(0) = [y,\psi,v]^{\textrm{T}}$, respectively. The optimal trajectory $\xi^{*} = [x,y,\psi,v]^{\textrm{T}}$ is generated by simulating the non-linear single-track vehicle model described in \cite{dixit2019trajectory} with the optimal inputs $u^{*}$ from the solution of MPC problem \eqref{eq:genOptProblem}. In the current implementation, it is assumed that the SV can perfectly track this reference trajectory.

\section{EVALUATION} \label{sec:results}
\subsection{Decision Model Performance} \label{sec:temporal}
\begin{figure}
	\centering
	\includegraphics[width=0.95\linewidth]{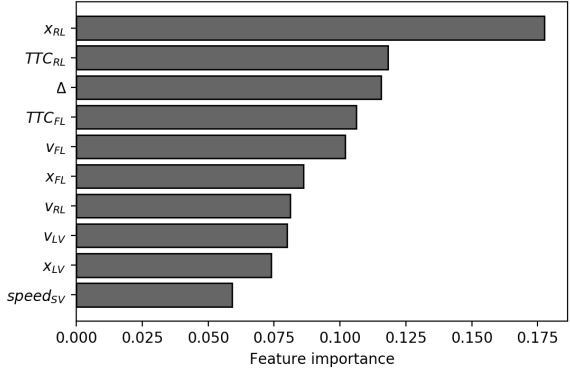}
	\caption{Feature importance values. The higher the value, the more important the feature is for classifying the manoeuvre decisions.}
	\label{fig:feature_importance}
\end{figure}

We assessed the decision model's performance on a withheld portion of the dataset consisting of 20\% of the available samples. We note that without the TTC and the utility features, as well as the exclusion of some lane changes from the dataset (see \Cref{sec:features}), the decision model's accuracy did not reach beyond 67\%. Considering that humans’ driving behaviour is inherently fuzzy, such low classification accuracy is to be expected.
The accuracy gain of 17\% shows that the engineered features successfully capture contextual information about a driving scene that is important for human drivers' decision making. The importance values for all the used features are shown in \autoref{fig:feature_importance}. Feature $x_{RL}$ is the most important in predicting the manoeuvre class, followed by the $TTC_{_{RL}}$ and $\Delta$. Other features associated with cars on the right adjacent lane and the vehicles' types (i.e., passenger car, truck or a motorcycle) showed to have no noticeable influence on the model's performance and hence were removed from the feature vector.

\begin{figure} 
	\centering
	\includegraphics[width=0.95\linewidth]{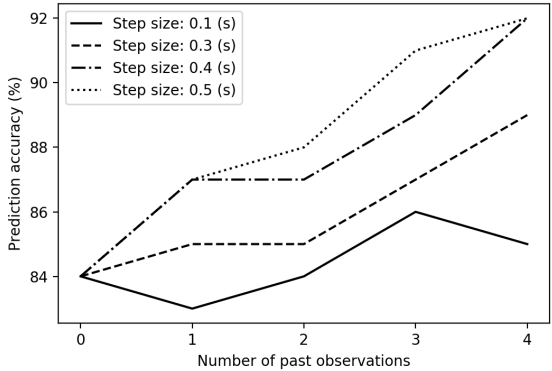}
	\caption{Model accuracy with inclusion of past vehicles' states in the feature vector.}
	\label{fig:history}
\end{figure}

We also studied the effect of including the past states of the ego vehicle and the cars surrounding it in the feature vector on the basis that past states may hold additional information such as cues about the interactions between the drivers in the scene. We experimented with a different number of past states and varied the length of the time-gap between them. 
The accuracy of the trained decision models is visualized in \autoref{fig:history}. The plot shows that the performance of the classifier improves as the number of past states is increased. Furthermore, the accuracy gain is greater for larger step-sizes. 

\subsection{Performance of the Integrated System}
\begin{figure}
	\centering
	\includegraphics[width=0.95\linewidth]{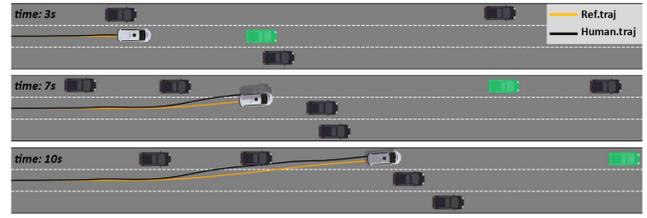}
	\caption{Development of an example traffic situation in which a lane change manoeuvre becomes desirable. The LV in the SV's desired lane is marked green. Both the ground truth trajectory and the simulated trajectory computed by MPC planner are shown for comparison.}
	\label{fig:example_scene}
\end{figure}

\begin{figure} 
	\centering
	\includegraphics[width=0.95\linewidth]{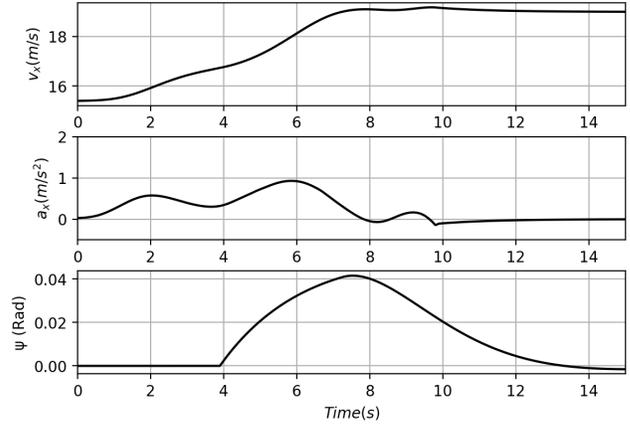}
	\caption{The longitudinal speed, acceleration action and steering action of the SV.}
	\label{fig:actions}
\end{figure}

As a qualitative performance measure of the approach, we performed simulations, whereby the actions of the simulated SV is logged while the dynamic environment evolves as per ground truth data recordings. This allows us to compare the trajectory that results from the actions taken by the SV to that of the human driver. 
A lane-change is only considered if its assigned probability is greater than a set threshold. The threshold defines the level of confidence required prior to initiating a lane change. In the current implementation, we have set the threshold to 80\%.
The computed trajectory of the simulated car and the trajectory of the human driver are shown in \autoref{fig:example_scene}. Acceleration and steering actions of the SV are shown in \autoref{fig:actions}. 
The result demonstrates a behaviour akin to that of the human driver. 
Another observation in this particular example is that the human driver completes the manoeuvre in a shorter duration, as indicated by the vehicle’s trajectory. However, this observation is not true across all the validation results.
To generate more human-like lane change trajectories, we attempted to predict the duration of a lane change for a given feature vector $\mathcal{F}$. However, we found there to be no correlation between the driving context and the lane change duration. This is presumably because the duration is primarily governed by the drivers' personal driving style rather than the immediate driving situation. 

\subsection{Sensitivity Analysis}
\begin{figure}
	\centering
	\includegraphics[width=0.98\linewidth]{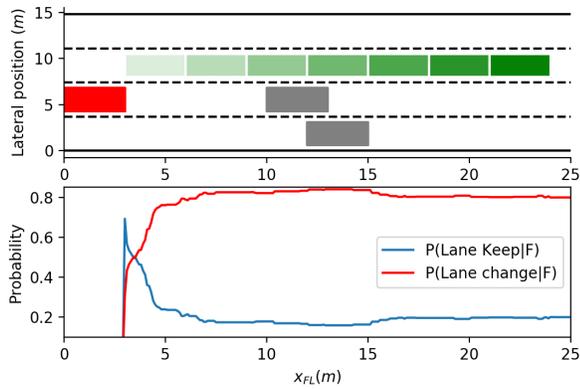}
	\caption{Probability assigned to each manoeuvre class for different positions of a vehicle (marked green) in SV's left adjacent lane.}
	\label{fig:sensitivity}
\end{figure}

We assessed how the probability assigned to each manoeuvre changes with the relative position of a vehicle in SV's desired lane. 
Value of $x_{FL}$ was manually changed from $\SI{0}{m}$ to $\SI{25}{m}$ while keeping the remaining feature values unchanged. The resulting probability values are plotted in \autoref{fig:sensitivity}.  
As expected, when the distance between the two vehicles is reduced, the assigned probability value for a lane change manoeuvre decreases. This relationship between the gap size and the manoeuvre class has emerged from the training data. We performed similar analysis with other features, but for the sake of brevity, the results are not included in this paper. We found that when the absolute value of the utility feature is low, the model becomes less sensitive to other features. Ambiguity among manoeuvre classes remains high regardless of the variation in other features' values. With a decision threshold of 80\%, many otherwise acceptable lane change instances would be missed, resulting in decisions that may not perfectly reflect the preferences of passengers onboard. Lowering the decision threshold is one solution, but doing so will result in decisions that are further from the nominal (i.e. commonly occurring) driving behaviour. 
	
\section{Conclusion and Future wok}\label{sec:conclusion}
In this paper, we presented an integrated decision-making and motion planning approach for highly automated driving on freeways. 
By making use of a set of features that were extracted from driving data, it is demonstrated that useful contextual information can be captured for generating driving behaviours akin to that of humans. Additionally, the planned trajectories are persistently feasible, providing a safety-net to the autonomous driving functionality. 

As a scope limitation for this paper, we assumed that a single decision model is representative of passengers' preferences. However, to achieve a more personalised driving style, a potential future direction is to cluster the drivers’ behaviour into different styles and train a decision model on each of the clusters \cite{morton2017simultaneous}.
Future work will also involve extending the current framework with a motion prediction module, such that the autonomous vehicle can predict other drivers' intentions and adapt its driving strategy accordingly.

\bibliographystyle{IEEEtran}
\bibliography{bib_collection, My_Collection_online, bibliography_SD}

\vspace{12pt}
\color{red}

\end{document}